\documentclass[dvips]{imsart}

\RequirePackage[OT1]{fontenc}
\RequirePackage{amsthm,amsmath}
\RequirePackage{natbib}
\RequirePackage[colorlinks,citecolor=blue,urlcolor=blue]{hyperref}

\startlocaldefs
\numberwithin{equation}{section}
\theoremstyle{plain}

\endlocaldefs

\usepackage[margin=0.9in]{geometry}

\usepackage[11pt]{moresize}
\usepackage{anyfontsize}
\usepackage{subfigure}
\usepackage{mathrsfs}
\usepackage{amssymb}
\usepackage{epsfig}
\usepackage{url}
\usepackage{natbib}
\usepackage{amsthm}
\usepackage{booktabs}
\usepackage{url}
\usepackage{enumerate}
\usepackage{multirow}
\usepackage{bm}
\usepackage{color}
\usepackage{algpseudocode}
\usepackage{algorithm}

\newcommand{\rx}{\mathtt{x}}

\newcommand{\ry}{\mathrm{y}}
\newcommand{\vx}{\mathbf{x}}
\newcommand{\vy}{\mathbf{y}}
\newcommand{\vz}{\mathbf{z}}
\newcommand{\base}{\mathit{h}}

\newcommand{\sD}{\mathscr{D}}

\newcommand{\sY}{\mathscr{Y}}

\newcommand{\sign}{\mathsf{sign}}

\newcommand{\nmathbf}{\bm}

\def\bfgamma  {\nmathbf \gamma}

\def\bfmu     {\nmathbf \mu}

\def\bfpi     {\nmathbf \pi}

\newcommand{\bfone}{{\nmathbf 1}}

\def\boldfacefake#1{\kern-4pt
    \hbox{ \mathsurround=0pt
    \hbox to 0.4pt{$#1$\hss}\hbox to 0.4pt{$#1$\hss}\hbox {$#1$}}}

\newcommand{\be}{\begin{eqnarray}}
\newcommand{\ee}{\end{eqnarray}}
\newcommand{\ba}{\begin{eqnarray*}}
\newcommand{\ea}{\end{eqnarray*}}

\newcommand{\reals}{\mbox{\rm I\kern-.20em R}}
\newcommand{\sreals}{\mbox{\small \rm I\kern-.20em R}}

\newtheorem{theorem0}{Theorem}
\newtheorem{lemma0}{Lemma}
\newtheorem{remark0}{Remark}
\newtheorem{fact0}{Fact}
\newtheorem{example0}{Example}
\newtheorem{definition0}{Definition}
\newtheorem{corollary0}{Corollary}
\newtheorem{proposition0}{Proposition}
\newtheorem{algorithmY}{Algorithm}
\newtheorem{conjecture0}{Conjecture}

\begin{document}

\begin{frontmatter}
\title{Prediction Error Reduction Function as a \\ Variable Importance Score}
%\thankstext{T1}{Footnote to the title with the ``thankstext'' command.}
\runtitle{PERF Score}

\begin{aug}
\author{\fnms{Ernest} \snm{Fokou\'e}\thanksref{t1,m1}\ead[label=e1]{epfeqa@rit.edu}}
\thankstext{t1}{Corresponding author}
%\thankstext{t2}{Associate Professor of Statistics}
\runauthor{Ernest Fokou\'e}

\affiliation{Rochester Institute of Technology\thanksmark{m1}}

\address{\thanksmark{m1}School of Mathematical Sciences\\
Rochester Institute of Technology\\
98 Lomb Memorial Drive, Rochester, NY 14623, USA\\
\printead{e1}}
\end{aug}

\begin{abstract}
\noindent This paper introduces and develops a novel variable importance score function
in the context of ensemble learning and demonstrates its appeal both
theoretically and empirically. Our proposed score function is simple and more straightforward than its counterpart proposed in the
context of random forest, and by avoiding permutations, it is by design computationally more efficient
than the random forest variable importance function. Just like the random forest variable importance function,
our score handles both regression and classification seamlessly. One of the distinct advantage of our
proposed score is the fact that it offers a natural cut off at zero, with all the positive scores indicating
importance and significance, while the negative scores are deemed indications of insignificance.
An extra advantage of our proposed score lies in the fact it works very well beyond ensemble of trees and can seamlessly be used with
any base learners in the random subspace learning context. Our examples, both simulated and real, demonstrate
that our proposed score does compete mostly favorably with the random forest score.
\end{abstract}


%\tableofcontents
\end{frontmatter}

\section{Introduction}
\noindent Consider a data set  $\mathscr{D}=\{ (\vx_1,\vy_1), \cdots,(\vx_n, \vy_n )\}$ where $\vx_i$ is a $p$-dimensional vector of attributes
of potentially different types observable on some input space denoted here by $\mathscr{X}$,
and  $\vy_i$ are the responses taken from $\mathscr{Y}$. We shall consider various scenarios, but mainly the regression scenario with $\mathscr{Y}=\mathbb{R}$
and the classification scenario with $\mathscr{Y}=\{1,2,\cdots,K\}$.
We consider the task of building the estimator $\widehat{f}(\cdot)$ of the true but unknown underlying $f$, and seek to build $\widehat{f}(\cdot)$
such that the true error (generalization error) is as small as possible. In this context, we shall use the
average test error  $\mathtt{AVTE}(\cdot)$, as our measure of predictive performance, namely
\begin{eqnarray}
    \label{eq:avte:1}
    \mathtt{AVTE}(\widehat{f}) =\frac{1}{R} \sum_{r=1}^{R} \left\{ \frac{1}{m} \sum_{j=1}^{m} \ell(\vy_{j}^{(r)}, \widehat{f}^{(r)}(\vx_j^{(r)}))\right\},
\end{eqnarray}
where  $\left(\vx_j^{(r)},\vy_j^{(r)}\right)$ is the $j$th observation from the test set at the $r$th random replication of the split of the data.
Throughout this paper, we shall use the zero-one loss \eqref{eq:loss:1} for all our classification tasks.
\begin{eqnarray}
    \label{eq:loss:1}
    \ell(\vy_{j}^{(r)}, \widehat{f}^{(r)}(\vx_j^{(r)})) = 1_{\{\vy_{j}^{(r)} \neq \widehat{f}^{(r)}(\vx_j^{(r)})\}}
                                           = \left\{\begin{array}{ll}
                                                 1 & \mbox{if $\vy_{j}^{(r)} \neq \widehat{f}^{(r)}(\vx_j^{(r)})$}\\
                                                 0 & \mbox{otherwise}.
                                                     \end{array}\right.
\end{eqnarray}
For regression tasks, we shall use the squared error loss \eqref{eq:loss:1}, namely
\begin{eqnarray}
    \label{eq:loss:2}
    \ell(\vy_{j}^{(r)}, \widehat{f}^{(r)}(\vx_j^{(r)}))=(\vy_{j}^{(r)} - \widehat{f}^{(r)}(\vx_j^{(r)}))^2.
\end{eqnarray}

Besides, seeking the optimal predictive estimator of $f$, we also seek to select the most important (useful) predictor variables as a byproduct of our overall learning
scheme. Indeed, while accurate prediction is very important in and of itself, it's often desirable or even crucial in some cases, provide
the added description of the importance of the variables involved in the prediction task. The statistical literature is filled with thousands of papers
on variable selection and measurement of variable importance.

\section{Main result}
We consider a framework with a $p$-dimensional input space $\mathscr{X}$ with typical input vector
$\vx =(\rx_1,\cdots, \rx_p)^\top$. We also consider building different models with different
subsets of the $p$ original variables. Let $\bfgamma = (\gamma_1, \cdots, \gamma_p)^{\top}$
denote the $p$-dimensional indicator such that
\begin{eqnarray}
\label{eq:indicator}
\gamma_{j} = \left\{\begin{array}{ll}
1 & \mbox{if $\rx_{j}$  is active in the current model indexed by $\bfgamma$}\\
0 & \mbox{otherwise}.
\end{array}\right.
\end{eqnarray}

Assume that we are given an ensemble (collection or aggregation) of models, say
\begin{eqnarray}
\label{eq:ensemble}
\mathscr{H} = \{\base(\cdot, \bfgamma^{(1)})), \base(\cdot, \bfgamma^{(2)})), \cdots, \base(\cdot, \bfgamma^{(B)}))\}
\end{eqnarray}
where $\base(\cdot, \bfgamma^{(b)}))$ denotes the function built with only those variables that are active in the $b$th model of the ensemble (aggregation), and
$\bfgamma^{(b)} = (\gamma_{1}^{(b)}, \cdots, \gamma_{p}^{(b)})$ with
\begin{eqnarray}
\label{eq:1:3}
\gamma_{j}^{(b)} = \left\{\begin{array}{ll}
1 & \mbox{if $\rx_{j}$  is active in the $b$-th model of the ensemble}\\
0 & \mbox{otherwise}.
\end{array}\right.
\end{eqnarray}

For instance, we may consider a homogeneous ensemble, i.e, an ensemble in which all the functions are of the same family, like the case where
all the base learners are multiple linear regression (MLR) models differing by the variables upon which they are built.
Consider a score function ${\tt score}(\base(\cdot, \bfgamma^{(b)}))$ used to assess the performance of model indexed by the variables active in $\bfgamma^{(b)}$.  We propose
a variable importance score in the form of a function that measures the importance of a variable $\rx_j$ in terms of the reduction in average score
\begin{eqnarray}
    \label{eq:1:2}
    {\tt PERF}(\rx_j) = \frac{1}{B}\sum_{b=1}^B{{\tt score}(\base(\cdot, \bfgamma^{(b)}))}-\frac{1}{B_j}\sum_{b=1}^B{\gamma_j^{(b)}{\tt score}(\base(\cdot, \bfgamma^{(b)}))}
\end{eqnarray}
where $B_j$ is the number of models containing the variable $\rx_j$, specifically $B_j = \sum_{b=1}^B{1_{\{\gamma_{j}^{(b)}=1\}}}$.
In words,
$$
{\tt PERF}(\rx_j) = \texttt{Average score over all models}-\texttt{Average score over all models with $\rx_j$}
$$
Intuitively, ${\tt PERF}(\rx_j)$ somewhat measures the impact of variable $\rx_j$. In the way similar to the approach used by sports writers to decide
the MVP on a team or in a league, ${\tt PERF}(\rx_j)$ looks at the overall performance of the whole ensemble
and then for each variable $\rx_j$ computes the direction and magnitude of the change to that overall performance of the ensemble brought by its presence in models. {\it If a variable $\rx_j$ is important, then its presence in any model will cause that model to perform better in the sense of having a lower than common average error (score). The average score of all models containing an important variable $\rx_j$ should therefore be lower than the overall average score.}
\begin{itemize}
\item $|{\tt PERF}(\rx_j)|$ measures the magnitude of the importance/impact.
\item $\sign({\tt PERF}(\rx_j))$ measures the direction of the impact.
\item If $\sign({\tt PERF}(\rx_j)) = +1$ and $|{\tt PERF}(\rx_j)|$ is relatively large, then $\rx_j$ is an important variable.
\end{itemize}
%
%
%$$
%\Pr[A] = \sum_{j=1}^{k}{\Pr[A \cap B_j]} = \Pr[A \cap B_1]+\Pr[A \cap B_2]+\cdots+\Pr[A \cap B_k]
%$$

\begin{itemize}
\item Seamlessly applied to large p small $n$.
\item All variables with ${\tt PERF}(\rx_j) \leq 0$ are unimportant and can be discarded.
\item The ${\tt PERF}(\cdot)$ score can be used whenever an ensemble $\mathscr{H}$ is available along
with a suitable score function for each base learner.
\item This works with any base learner and can be adapted to parametric, nonparametric and semi-parametric models
and one can imagine ensembles with any base learners as its atoms.
%\item Other score functions for the base learners are
\item A great advantage over the traditional variable importance \cite{Breiman:2001:1}, \cite{Breiman:2001:2} 
score functions is that the clear cut-off at zero,
in the sense that all variables with ${\tt PERF}(\rx_j) > 0$ are kept and all those variables
with ${\tt PERF}(\rx_j) \leq 0$ are thrown away.
\end{itemize}

\subsection{PERF score via Random Subspace Learning}
A natural implementation of ${\tt PERF}(\cdot)$ can be done using the ubiquitous bootstrap along with
the random subspace learning scheme. The {\tt Out-of-Bag (oob)} error  in the bagging or random subspace learning context is a good (in fact excellent)
candidate score function, especially when the goal if the selection of variables that lead to
the lowest prediction error. The advantage of using {\tt oob} as the score lies in the fact that the score
is obtained as part of building the ensemble in the random subspace learning framework.
Consider the training set $\sD = \{\vz_i=(\vx_i^\top,\ry_i)^\top,\,\,i=1,\cdots,n\}$, where $\vx_i^\top = (\rx_{i1},\cdots,\rx_{ip}) $ and  $\ry_i\in \sY$ are realizations of two random variables $X$ and $Y$ respectively. Let $\vx_{i,{\bfpi_j}} = (\rx_{i,1},\cdots,\rx_{i,\bfpi_j},\cdots,\rx_{i,d})$.
The permutation $\bfpi_j$ acts the $|\bar{\sD}^{(b)}|$-dimensional $j$th column of the out-of-bag data matrix.
Essentially, $\bfpi_j$ simply permutes the $|\bar{\sD}^{(b)}|$ elements of the $j$th column of the out-of-bag data matrix.

\begin{algorithm}
\caption{PERF Score Estimate via Random Subspace Learning}\label{algo:rf:1}
\begin{algorithmic}[1]
\Procedure{PERF Score}{$B$}\Comment{Computing the PERF Score based on $B$ base learners}
\State {\tt Choose a base learner $\widehat{\base}(\cdot)$} \Comment{e.g.:  Trees, MLR}
\State {\tt Choose an estimation method} \Comment{e.g.:  Recursive Partitioning or OLS}
\State {\tt Initialize all the $\widehat{\tt PERF}(\rx_j)$ and $\widehat{\tt VI}(\rx_j)$ at zero}
\For{$b=1$ to $B$}
\State {\tt Draw with replacement} from $\sD$ a bootstrap sample $\sD^{(b)} = \{\vz_1^{(b)},\cdots,\vz_n^{(b)}\}$
\State {\tt Draw without replacement} from $\{1,\cdots,p\}$ a subset $\mathscr{V}^{(b)}=\{j_1^{(b)},\cdots,j_d^{(b)}\}$ of $d$ variables.
\State {\tt Form the indicator vector}  $\bfgamma^{(b)} = (\gamma_{j}^{(b)}, \cdots, \gamma_{p}^{(b)})$ with
$$
\gamma_{j}^{(b)} = \left\{\begin{array}{ll}
1 & \mbox{if $j \in \{j_1^{(b)},\cdots,j_d^{(b)}\}$  }\\
0 & \mbox{otherwise}.
\end{array}\right.
$$
\State {\tt Drop unselected variables} from $\sD^{(b)}$ so that $\sD_{\tt sub}^{(b)}$ is $d$ dimensional
\State {\tt Build the $b$th base learner} $\widehat{\base}(\cdot, \bfgamma^{(b)})$ based on $\sD_{\tt sub}^{(b)}$
\State {\tt Compute score of the $b$th base learner} $\widehat{\base}(\cdot, \bfgamma^{(b)})$  \Comment{e.g. Out-of-bag error}
$$
{\tt s}^{(b)} = {\tt score}(\widehat{\base}(\cdot, \bfgamma^{(b)})) = \frac{1}{|\bar{\sD}^{(b)}|}\sum_{\vz_i \notin \sD^{(b)}}{\ell(\ry_i,\widehat{\base}(\vx_i, \bfgamma^{(b)}))}
$$
 \For{$j \in \mathscr{V}^{(b)}$}
 \State {\tt Generate the permutation of the $j$th column of $\bar{\sD}^{(b)}$, namely}
 $$
 {\bfpi_j}
 $$
 \State {\tt Compute the permutation impacted score}
$$
{\tt s}_{\bfpi_j}^{(b)} = {\tt score}_{\bfpi_j}(\widehat{\base}(\cdot, \bfgamma^{(b)})) = \frac{1}{|\bar{\sD}^{(b)}|}\sum_{\vz_i \notin \sD^{(b)}}{\ell(\ry_i,\widehat{\base}(\vx_{i,{\bfpi_j}}, \bfgamma^{(b)}))}
$$
 \State {\tt Compute the $b$th instance of the importance of $\rx_j$}
 $$
 \widehat{\tt VI}^{(b)}(\rx_j) = {\tt s}^{(b)} - {\tt s}_{\bfpi_j}^{(b)}
 $$
\EndFor \label{algo:rf:2}
\EndFor \label{algo:rf:1}
\State Use the ensemble $\mathscr{H} = \Big\{\widehat{\base}(\cdot, \bfgamma^{(b)}),\, b=1,\cdots, B\Big\}$ to form the estimator
\begin{eqnarray}
    \label{eq:perf:rssl:1}
    \widehat{\tt PERF}(\rx_j) = \frac{1}{B}\sum_{b=1}^B{{\tt score}(\widehat{\base}(\cdot, \bfgamma^{(b)}))} - \frac{1}{B_j}\sum_{b=1}^B{\gamma_j^{(b)}{\tt score}(\widehat{\base}(\cdot, \bfgamma^{(b)}))}
\end{eqnarray}
\begin{eqnarray}
    \label{eq:vi:rssl:1}
    \widehat{\tt VI}(\rx_j) = \frac{1}{B_j}\sum_{b=1}^B{\gamma_j^{(b)}\widehat{\tt VI}^{(b)}(\rx_j)}
\end{eqnarray}
\EndProcedure
\end{algorithmic}
\end{algorithm}

\section{Computational demonstrations}
\subsection{Simulated Example}
The dataset in this example is simulated data with different scenarios on the level of correlation among the variables, and the
ratio $n$ and $p$. In this particular example, the true function is
$$
f(\vx) = 1 + 2\rx_3 + \rx_7 + 3\rx_9
$$
with  $\vx \sim {\tt MVN}(\bfone_9, \Sigma_{\rho})$ and $\epsilon \sim {\bf N}(0,1)$.
The dataset in this example is simulated data with different scenarios on the level of correlation among the variables, and the
ratio $n$ and $p$. Specifically, we simulate data by defining $\rho \in [0,1)$, then we generate our predictor variables using a
multivariate normal distribution. Throughout this paper, the multivariate Gaussian density will be denoted by $\phi_p(\vx; \bfmu, \Sigma)$
\begin{eqnarray}
   \label{eq:p:gauss}
    \phi_p(\vx; \bfmu, \Sigma) = \frac{1}{\sqrt{(2\pi)^p |\Sigma|}}
    \exp\left\{-\frac{1}{2}(\vx-\bfmu)^\top \Sigma^{-1}(\vx-\bfmu)\right\}
\end{eqnarray}

Furthermore, in order to study the effect of the correlation pattern, we simulate the data using a covariance matrix $\Sigma$ parameterized by
$\tau$ and $\rho$ and defined by $\tau\Sigma$ where $\Sigma= (\sigma_{ij})$ with $\sigma_{ij}=\rho^{|i-j|}$.
$$\Sigma = \Sigma(\tau, \rho) = \tau\left(
                        \begin{array}{ccccc}
                          1 & \rho & \cdots & \rho^{p-2} &\rho^{p-1} \\
                          \rho & 1 & \rho & \cdots & \rho^{p-2} \\
                          \vdots & \ddots & \ddots & \ddots &  \vdots\\
                          \rho^{p-2} & \ddots & \rho & 1 &  \rho\\
                          \rho^{p-1} & \rho^{p-2} & \cdots & \rho & 1 \\
                        \end{array}
                      \right)
$$
For simplicity however, we use the first $\Sigma$ with $\tau=1$ throughout this paper. For the remaining
parameters, we use $\rho \in \{0, 0.25, 0.75\}$ and $p \in \{17, 250\}$, with the same $n=200$.
%
%The Prestige data set has  $n=102$ and $p=4$.
%
%Another dataset used in our comparison is the prostate cancer data set which is rather small in size compared to the previous
%dataset with $n=97$ and $p=8$.
%
%The cars data set has  $n=80$ and $p=4$. Here we seek to predict the gas mileage (MPG) of
%a car given four characteristics, namely, VOL, HP, SP and WT.
%
%The Boston housing data set has  $n=506$ and $p=13$.
%

\begin{figure}
\begin{center}
\subfigure[Permutation-free Variable Importance. \label{fig:simulated-high-corr-ldhss-perf-1a}]{
\resizebox*{6.5cm}{!}{\includegraphics[width=7.5cm, height=7.5cm, angle=270]{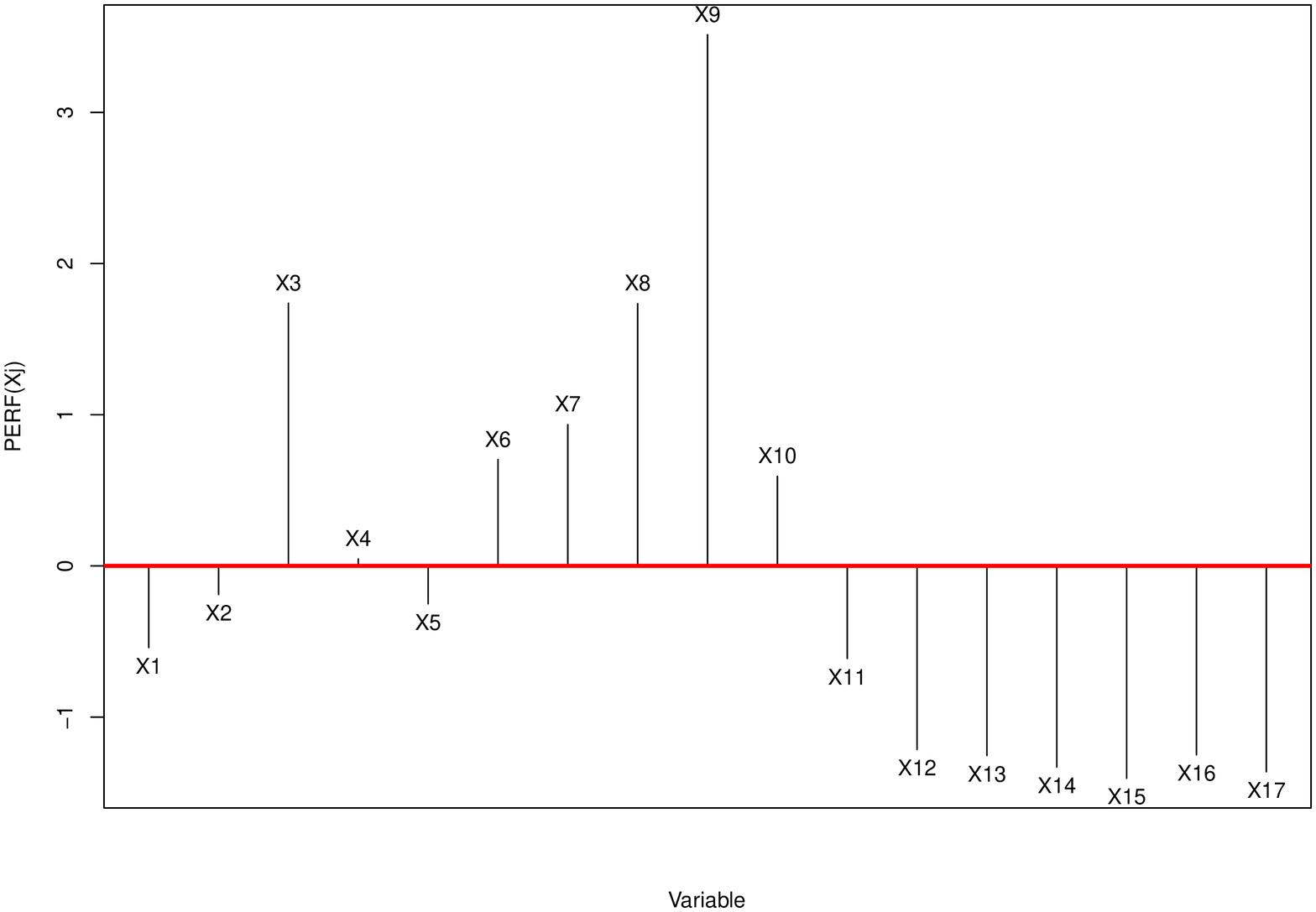}}}\hspace{6pt}
\subfigure[Permutation-based Variable Importance.\label{fig:simulated-high-corr-ldhss-perf-1b}]{
\resizebox*{6.5cm}{!}{\includegraphics[width=7.5cm, height=7.5cm, angle=270]{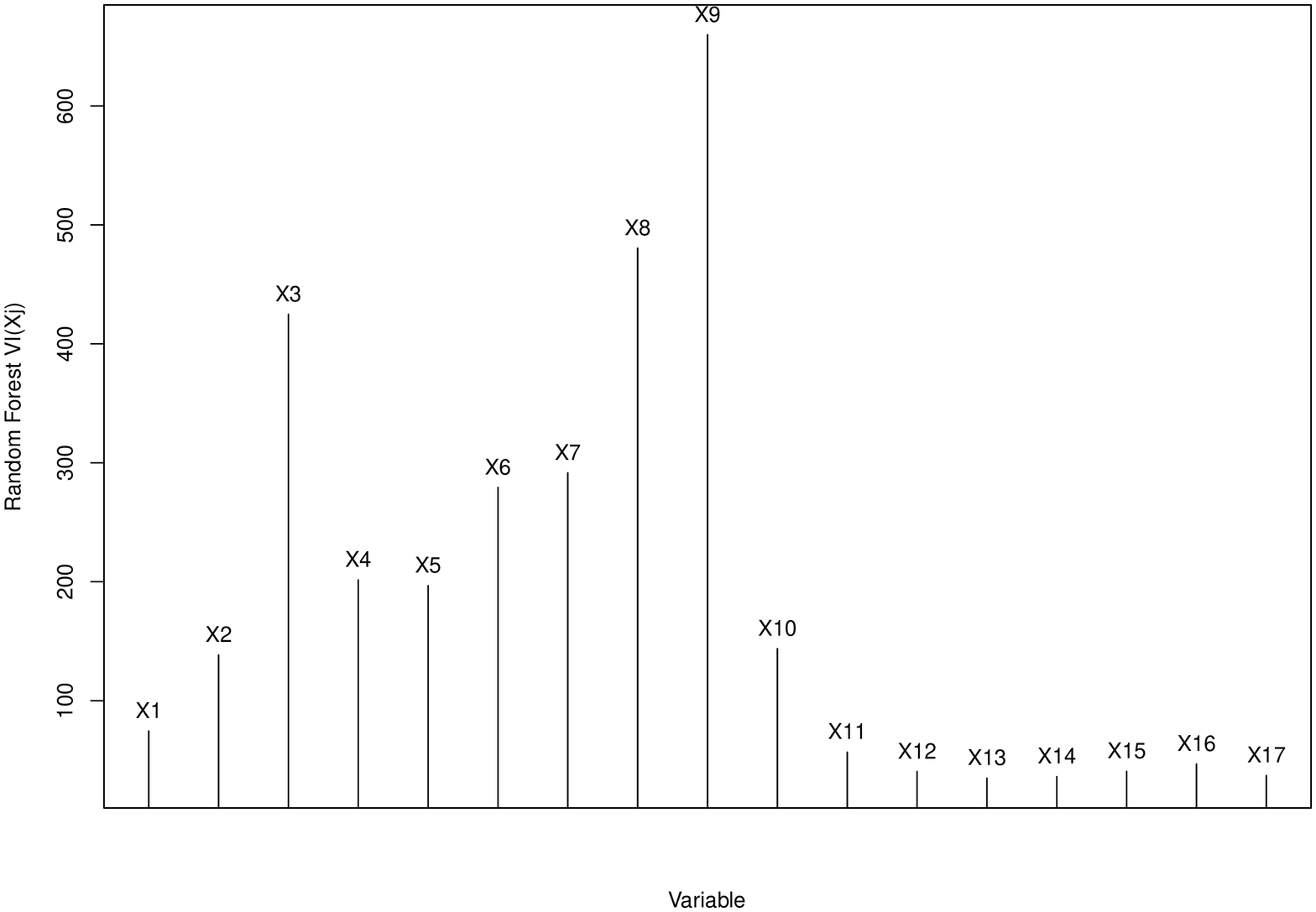}}}
 \caption{Variable score for simulated data with high correlation among the variables in low dimension high sample size setting}
 \label{fig:simulated-high-corr-ldhss-perf-1}
\end{center}
\end{figure}

\begin{figure}
\begin{center}
\subfigure[Permutation-free Variable Importance. \label{fig:simulated-mild-corr-ldhss-perf-1a}]{
\resizebox*{6.5cm}{!}{\includegraphics[width=7.5cm, height=7.5cm, angle=270]{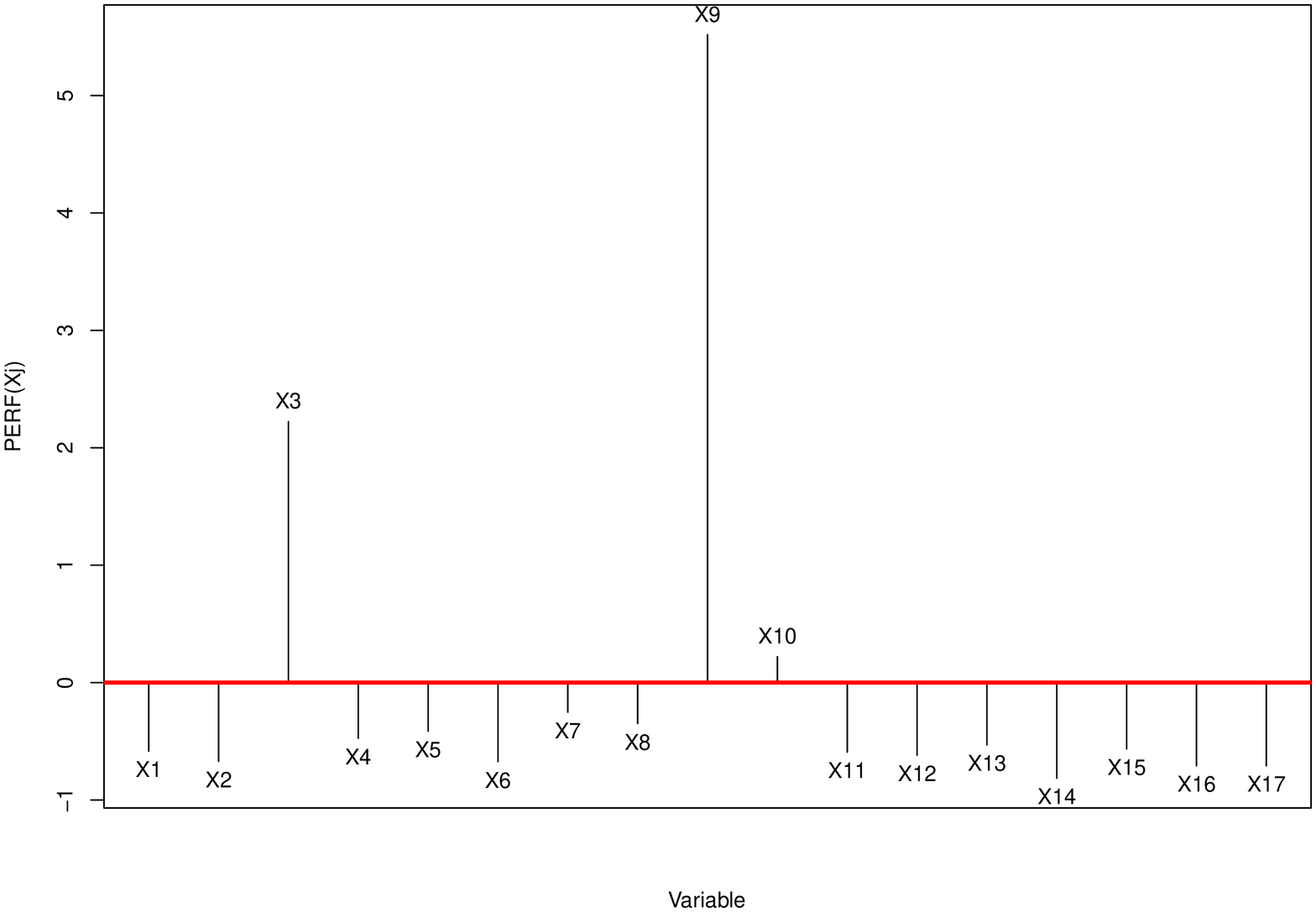}}}\hspace{6pt}
\subfigure[Permutation-based Variable Importance.\label{fig:simulated-mild-corr-ldhss-perf-1b}]{
\resizebox*{6.5cm}{!}{\includegraphics[width=7.5cm, height=7.5cm, angle=270]{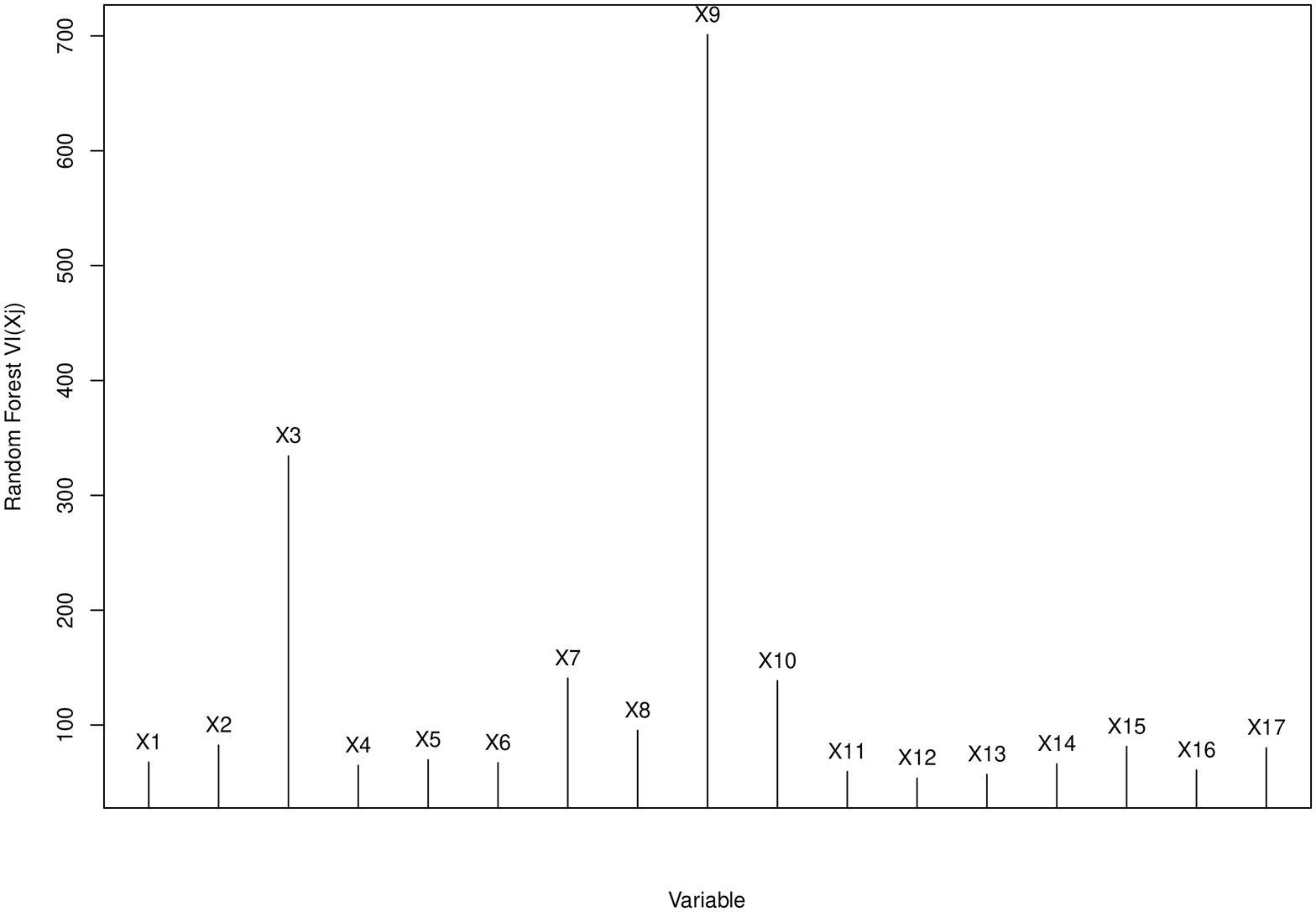}}}
 \caption{Variable Importance Scores for simulated data with mild correlation among the variables in low dimension high sample size setting}
 \label{fig:simulated-mild-corr-ldhss-perf-1}
\end{center}
\end{figure}

\begin{figure}
\begin{center}
\subfigure[Permutation-free Variable Importance. \label{fig:simulated-zero-corr-ldhss-perf-1a}]{
\resizebox*{6.5cm}{!}{\includegraphics[width=7.5cm, height=7.5cm, angle=270]{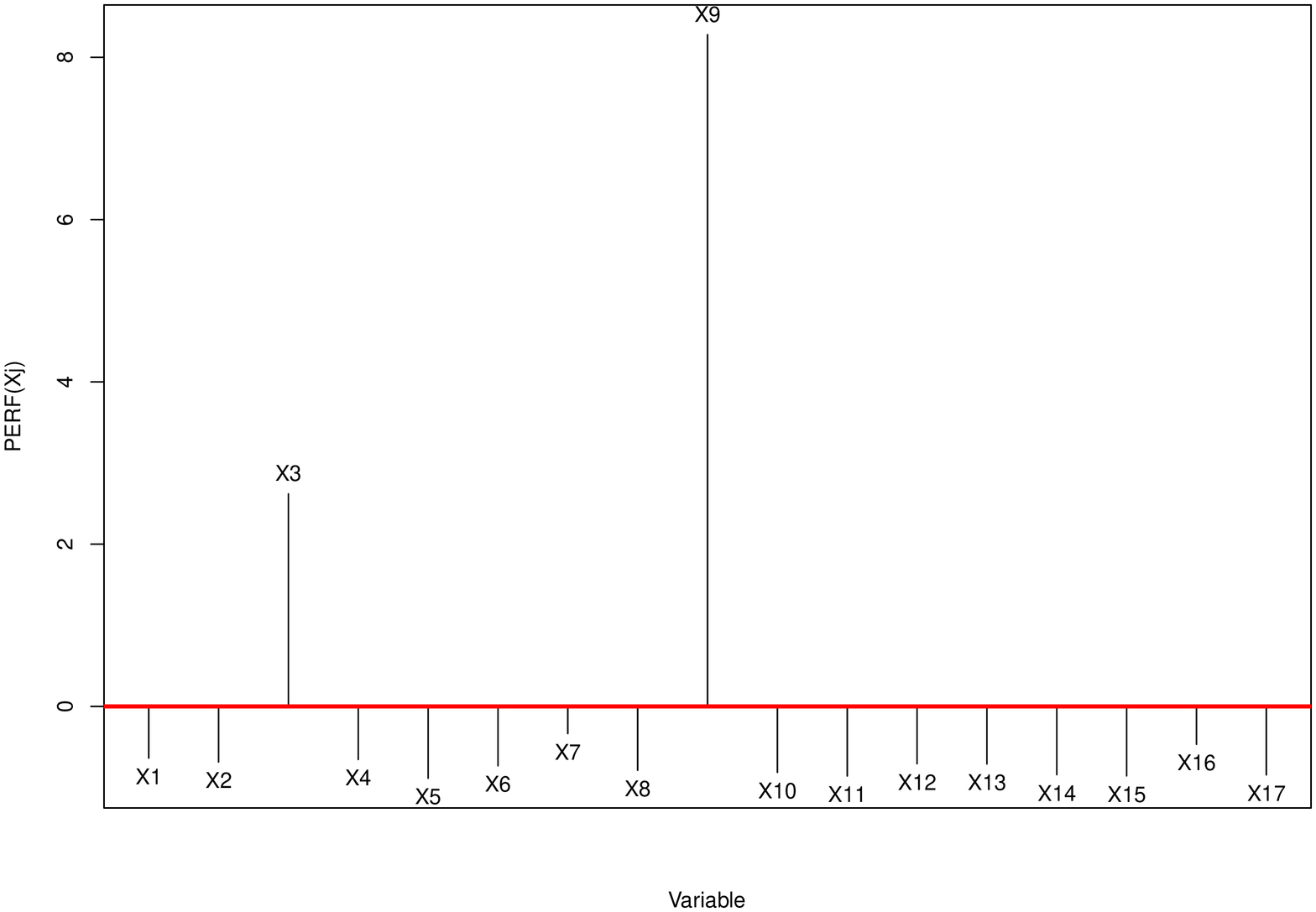}}}\hspace{6pt}
\subfigure[Permutation-based Variable Importance.\label{fig:simulated-zero-corr-ldhss-perf-1b}]{
\resizebox*{6.5cm}{!}{\includegraphics[width=7.5cm, height=7.5cm, angle=270]{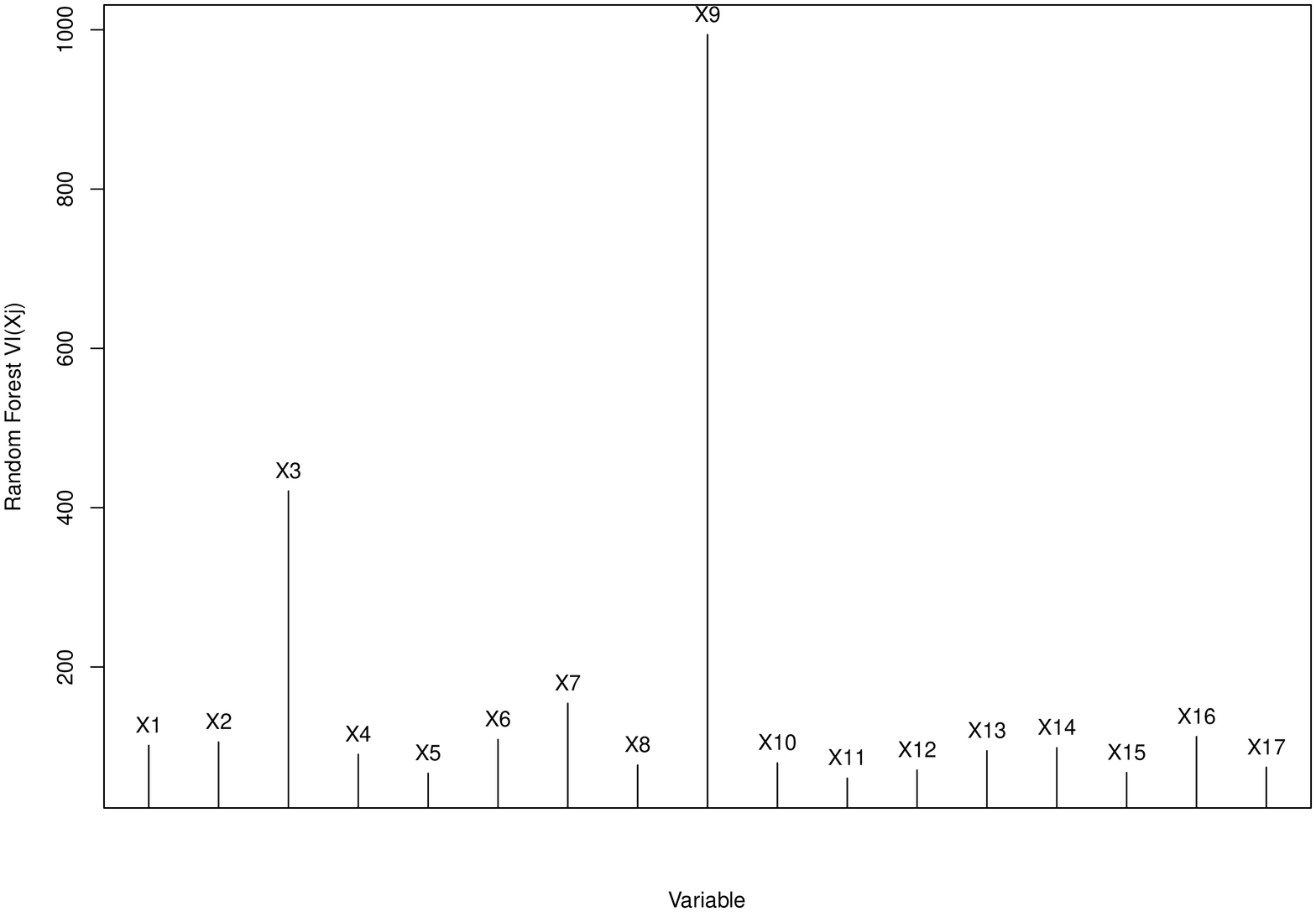}}}
 \caption{Variable Importance Scores for simulated data with zero correlation among the variables in low dimension high sample size setting}
 \label{fig:simulated-zero-corr-ldhss-perf-1}
\end{center}
\end{figure}

\begin{figure}
\begin{center}
\subfigure[Permutation-free Variable Importance. \label{fig:attitude-perf-1a}]{
\resizebox*{6.5cm}{!}{\includegraphics[width=7.5cm, height=7.5cm, angle=270]{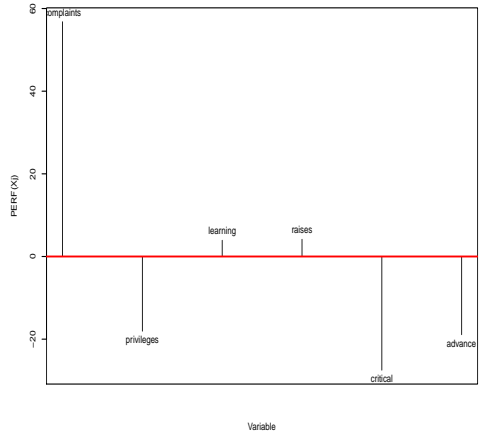}}}\hspace{6pt}
\subfigure[Permutation-based Variable Importance.\label{fig:attitude-perf-1b}]{
\resizebox*{6.5cm}{!}{\includegraphics[width=7.5cm, height=7.5cm, angle=270]{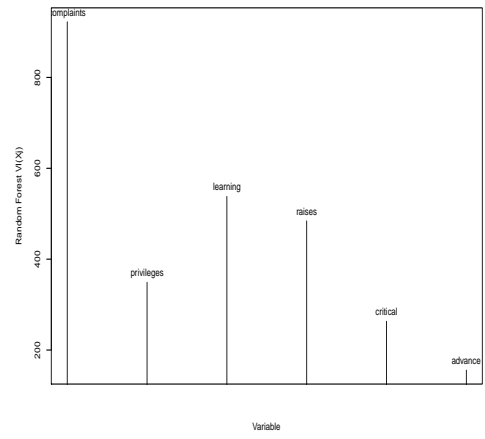}}}
 \caption{Variable Importance Scores for the Attitude Data Set, for which $n=30$ and $p=6$.}
 \label{fig:attitude-perf-1}
\end{center}
\end{figure}

%\begin{figure}
%  \centering
%  % Requires \usepackage{graphicx}
%  \includegraphics[width=7.5cm, height=7.5cm, angle=270]{cars-perf-1}
%  \includegraphics[width=7.5cm, height=7.5cm, angle=270]{cars-rf-1}
%  \caption{PERF score for simulated data with zero correlation among the variables in low dimension high sample size setting}
%  \label{fig:cars-perf-1}
%\end{figure}

%
%\begin{figure}
%  \centering
%  % Requires \usepackage{graphicx}
%  \includegraphics[width=7.5cm, height=7.5cm, angle=270]{bodyfat-perf-1}
%  \includegraphics[width=7.5cm, height=7.5cm, angle=270]{bodyfat-rf-1}
%  \caption{PERF score for simulated data with zero correlation among the variables in low dimension high sample size setting}
%  \label{fig:bodyfat-perf-1}
%\end{figure}
%
%\begin{figure}
%  \centering
%  % Requires \usepackage{graphicx}
%  \includegraphics[width=7.5cm, height=7.5cm, angle=270]{bostonhousing-perf-1}
%  \includegraphics[width=7.5cm, height=7.5cm, angle=270]{bostonhousing-rf-1}
%  \caption{PERF score for simulated data with zero correlation among the variables in low dimension high sample size setting}
%  \label{fig:bostonhousing-perf-1}
%\end{figure}

%
%\begin{figure}
%  \centering
%  % Requires \usepackage{graphicx}
%  \includegraphics[width=7.5cm, height=7.5cm, angle=270]{prostate-perf-1}
%  \includegraphics[width=7.5cm, height=7.5cm, angle=270]{prostate-rf-1}
%  \caption{PERF score for simulated data with zero correlation among the variables in low dimension high sample size setting}
%  \label{fig:prostate-perf-1}
%\end{figure}

\begin{figure}
\begin{center}
\subfigure[Permutation-free Variable Importance. \label{fig:pima-perf-1a}]{
\resizebox*{6.5cm}{!}{\includegraphics[width=7.5cm, height=7.5cm, angle=270]{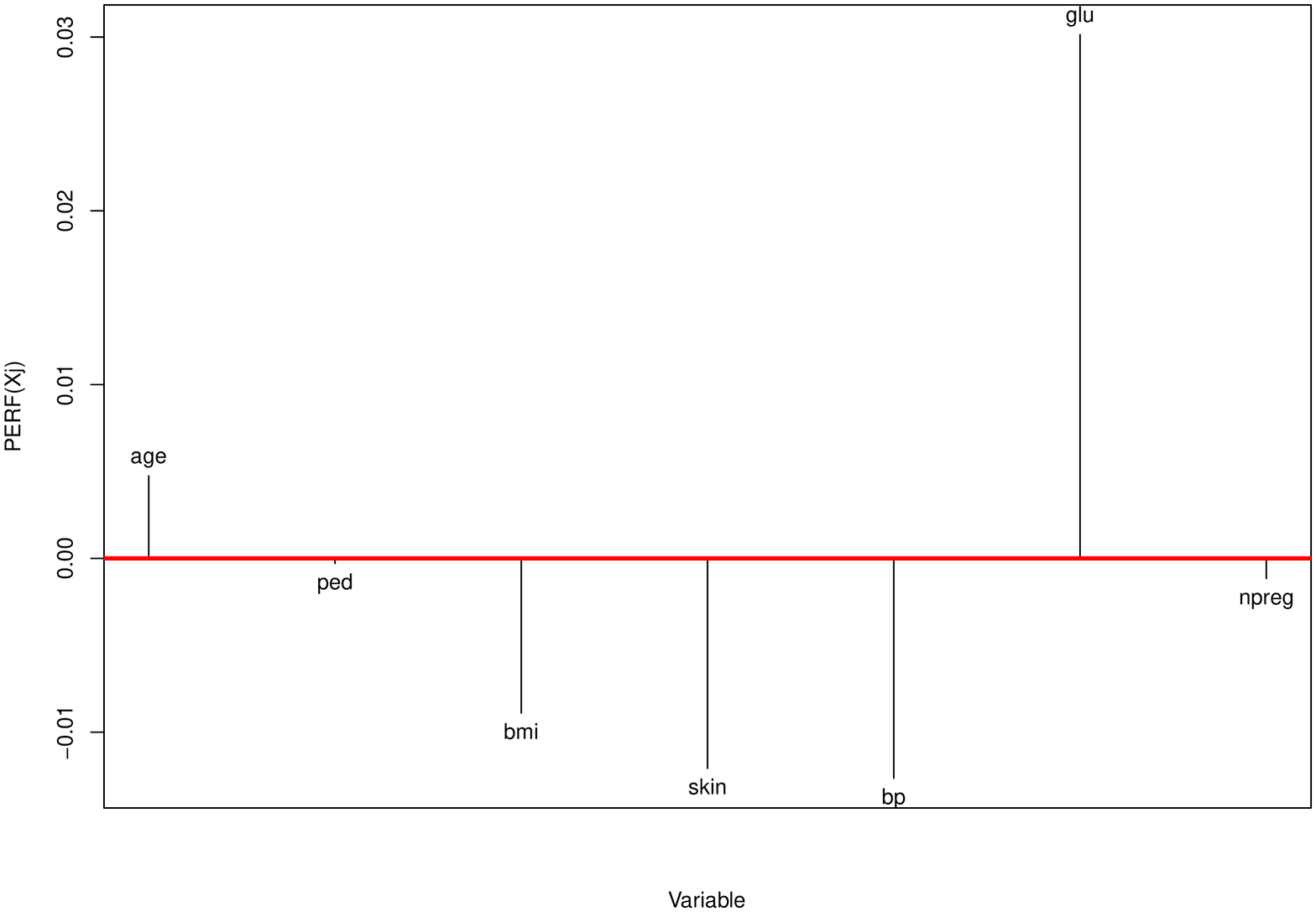}}}\hspace{6pt}
\subfigure[Permutation-based Variable Importance.\label{fig:pima-perf-1b}]{
\resizebox*{6.5cm}{!}{\includegraphics[width=7.5cm, height=7.5cm, angle=270]{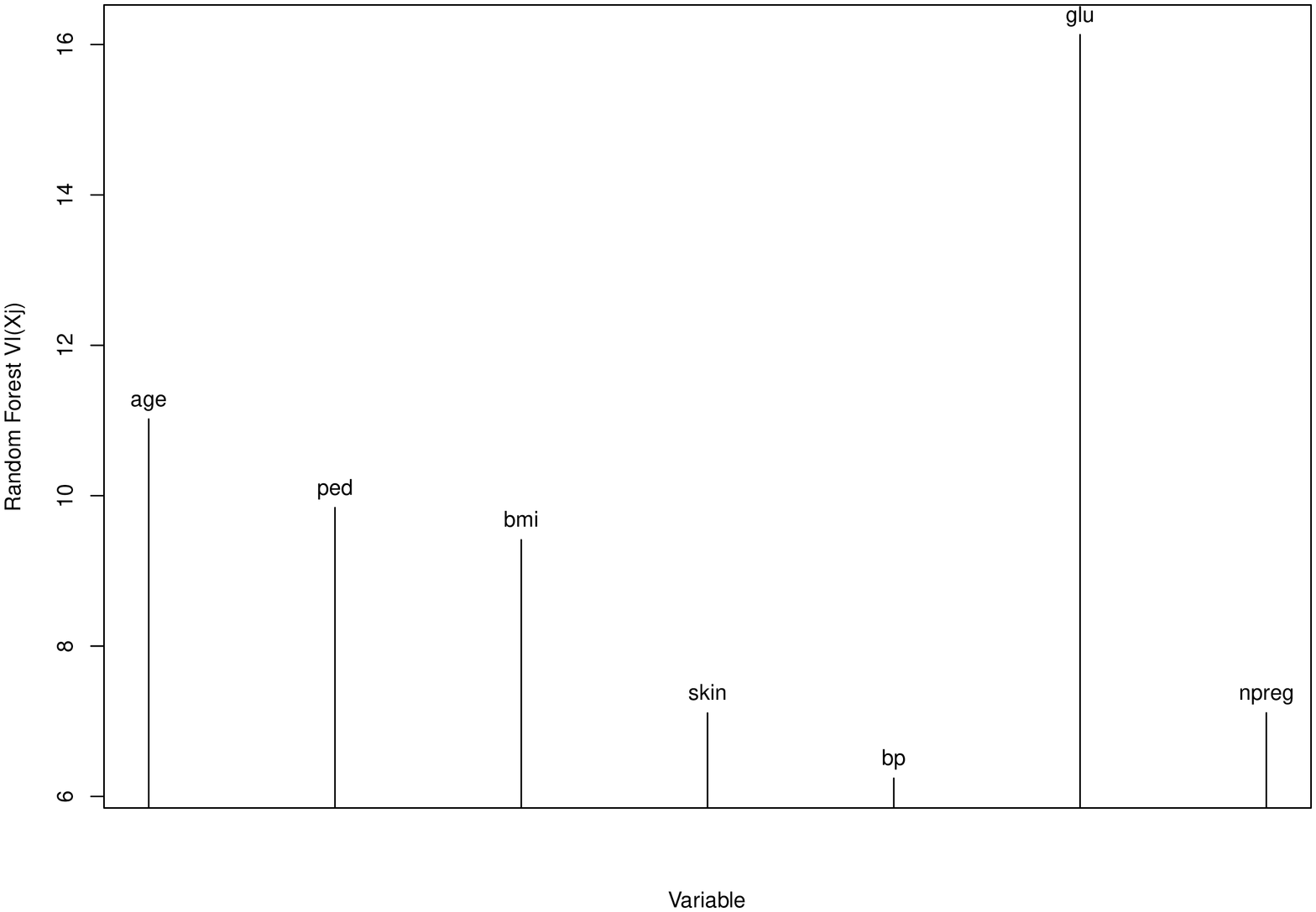}}}
 \caption{Variable Importance Scores for the Spam Detection Dataset where $n=200$ and $p=7$, and $K=2$ classes.}
 \label{fig:pima-perf-1}
\end{center}
\end{figure}

\begin{figure}
\begin{center}
\subfigure[Permutation-free Variable Importance. \label{fig:spam-perf-1a}]{
\resizebox*{6.5cm}{!}{\includegraphics[width=7.5cm, height=7.5cm]{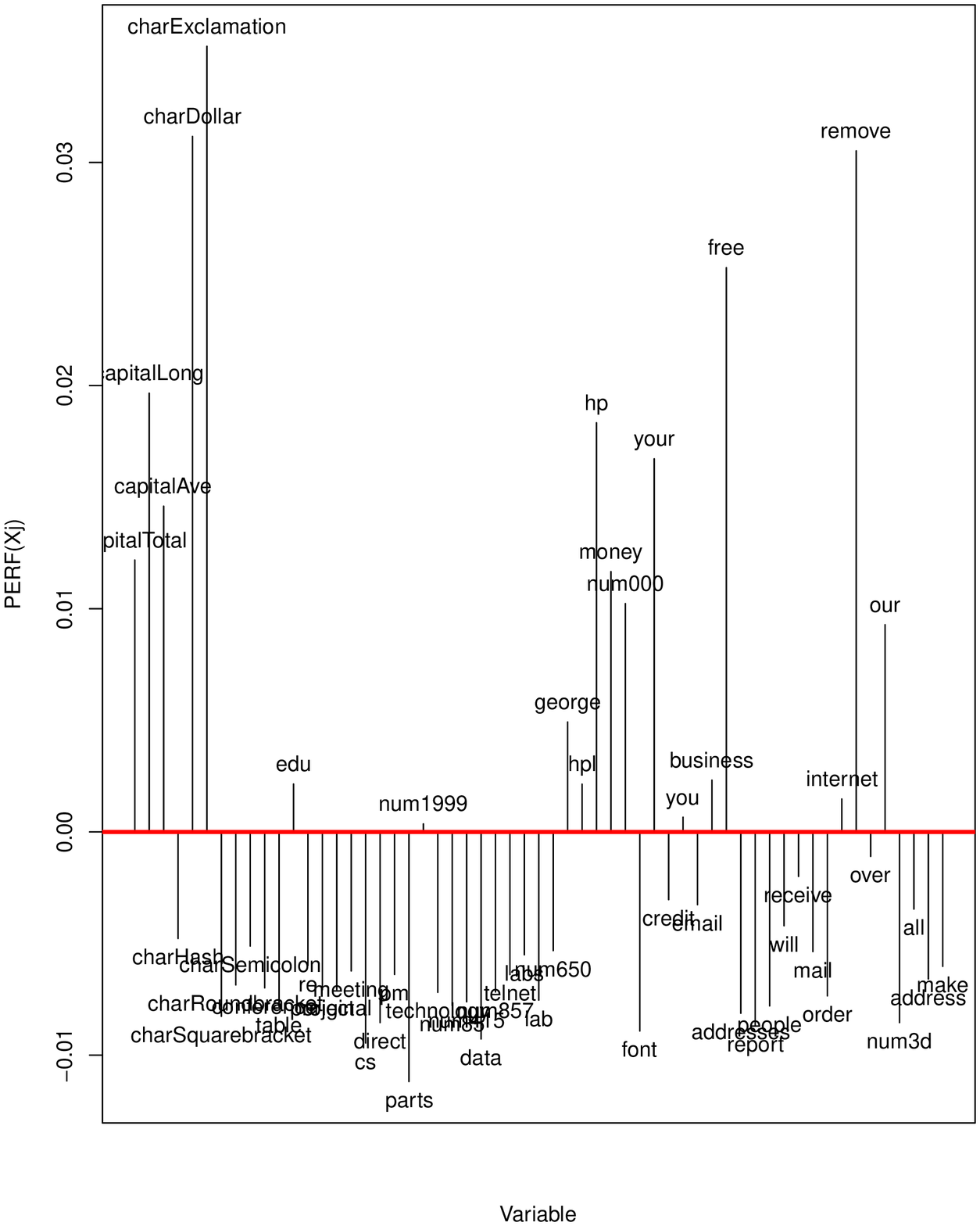}}}\hspace{6pt}
\subfigure[Permutation-based Variable Importance.\label{fig:spam-perf-1b}]{
\resizebox*{6.5cm}{!}{\includegraphics[width=7.5cm, height=7.5cm]{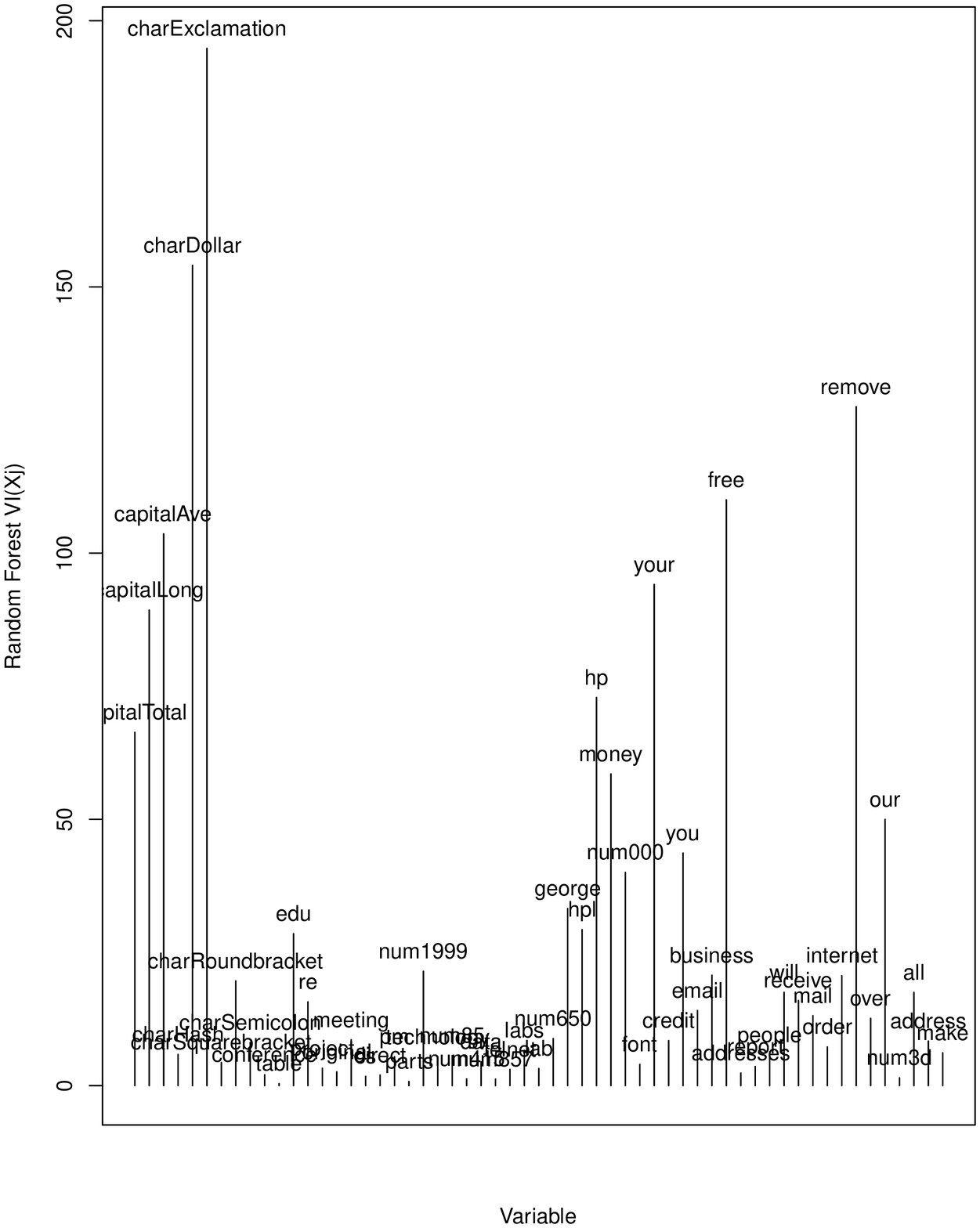}}}
 \caption{Variable Importance Scores for the Spam Detection Dataset where $n=4601$ and $p=57$, and $K=2$ classes.}
 \label{fig:spam-perf-1}
\end{center}
\end{figure}

\section{Conclusion and Discussion}
We have presented a variable importance score function in the context of ensemble learning.
Our proposed score function is simple and more straightforward than its counterpart proposed in the
context of random forest, and by avoiding permutations, it is by design computationally more efficient
than the random forest variable importance function. Just like the random forest variable importance function,
our score handles both regression and classification seamlessly. One of the distinct advantage of our 
proposed score is the fact that it offers a natural cut off at zero, with all the positive scores indicating
importance and significance, while the negative scores are deemed indications of insignificance. 
An extra advantage of our proposed
score lies in the fact it works very well beyond ensemble of trees and can seamlessly be used with
any base learners in the random subspace learning context. Our examples, both simulated and real, demonstrated 
that our proposed score does compete mostly favorably with the random forest score.
In our future work, we present and compare the corresponding average test errors of the single models
made up of the most important variables. We also provide in our future work theoretical proofs of 
the connection between our score function and the significance of variables selected using existing criteria.
It is also our plan to address the fact that sometimes the correlation structure among the predictor variables
obscures the ability of our proposed score to correctly identify some significant variables.

\section*{Acknowledgements}
Ernest Fokou\'e wishes to express his heartfelt gratitude and infinite thanks to Our Lady of Perpetual Help for Her
ever-present support and guidance, especially for the uninterrupted flow of inspiration received through Her
most powerful intercession.

\bibliographystyle{chicago}
\bibliography{perf-score-1}

\begin{thebibliography}{}

\bibitem[\protect\citeauthoryear{Breiman}{Breiman}{2001a}]{Breiman:2001:1}
Breiman, L. (2001a).
\newblock Random forests.
\newblock {\em Machine Learning\/}~{\em 45}, 5--32.

\bibitem[\protect\citeauthoryear{Breiman}{Breiman}{2001b}]{Breiman:2001:2}
Breiman, L. (2001b, August).
\newblock Statistical modeling: The two cultures.
\newblock {\em Statistical Science\/}~{\em 16\/}(3), 199--215.

\end{thebibliography}

%\bibliography{Tanujit-Ernest-Varanasi-1}
\end{document}